\documentclass[11pt]{amsart}

\usepackage{graphicx}
\usepackage{pifont}
\usepackage{amssymb,amsmath,amsfonts}

\newtheorem{lemma}{Lemme}

\begin{document}
\title{Choix d'un espace de repr\'esentation image adapt\'e \`a la d\'etection de r\'eseaux routiers}
\author{J\'er\^ome Gilles}
\address{
DGA/Centre d'Expertise Parisien\\
D\'epartement G\'eographie-Imagerie-Perception, \\
16 bis, rue Prieur de la C\^ote d'Or\\
94114 Arcueil Cedex, France\\    
T\`el: int+ 33 1 42 31 94 27, Fax: int+ 33 1 42 31 99 64\\
}
\email{jerome.gilles@etca.fr}

\begin{abstract}
    Ces derni\`eres ann\'ees, des algorithmes permettant de d\'ecom\-poser une image en ses composantes structures + textures ont vu le jour. Dans cet article, nous pr\'esentons une application de ce type de d\'ecomposition au probl\`eme de la d\'etection de r\'eseau routier en imagerie a\'erienne ou satellitaire. La cha\^ine algorithmique met en \oe uvre la d\'ecomposition d'image (dont nous utilisons une propri\'et\'e particuli\`ere), une d\'etection d'alignements issus de la th\'eorie de la \textit{Gestalt} et un raffinement des routes extraites par contours actifs statistiques.
\end{abstract}

\maketitle

\section{Introduction}
Diff\'erents travaux, plus ou moins r\'ecents, se sont attel\'es au probl\`eme de la d\'etection automatique de r\'eseaux routiers en imagerie a\'erienne ou satellitaire (\cite{geman,lacoste}). Ces travaux ont plus particuli\`erement concern\'e les m\'ethodes de d\'etection et la mod\'elisation m\^eme d'une route.\\
Dans cet article, nous \'etudions la possibilit\'e de disposer d'un espace de repr\'esentation de l'image qui soit mieux adapt\'e en vue de faire la d\'etection. Nous proposons d'utiliser l'espace de textures bas\'e sur les m\'ethodes de d\'ecomposition d'image d\'evelopp\'ees ces derni\`eres ann\'ees autour des travaux de Y.Meyer (\cite{meyer}). Apr\`es avoir rappel\'e le principe de la d\'ecomposition d'image, nous montrerons que la composante texture permet de r\'ehausser les objets filiformes. La d\'ecomposition sera alors utilis\'ee comme pr\'etraitement avant l'application d'un algorithme de d\'etection (nous proposons un algorithme bas niveau bas\'e sur la m\'ethode de d\'etection d'alignements issue de la th\'eorie de la \textit{Gestalt} et des contours actifs statistiques).

\section{D\'ecomposition d'image}
\subsection{Mod\`ele $u+v$}
Dans \cite{meyer}, Yves Meyer, se basant sur l'algorithme de Rudin-Osher-Fatemi \cite{rof}, propose un mod\`ele permettant de d\'ecomposer une image en deux parties: l'une (not\'ee $u$) contenant les structures, l'autre (not\'ee $v$) contenant les textures. Ce mod\`ele consiste \`a minimiser la fonctionnelle (\ref{eqn:eq1}):

\begin{equation}\label{eqn:eq1}
F^{YM}(u,v)=\|u\|_{BV}+\lambda\|v\|_G
\end{equation}

o\`u $u\in BV$ (l'espace des fonctions \`a variations born\'ees) et $v\in G$ (espace des fonctions oscillantes, proche du dual de $BV$, et ayant comme propri\'et\'e que plus une fonction est oscillante, plus sa norme $\|.\|_G$ sera faible), $\lambda$ \'etant un param\`etre du mod\`ele. Ce mod\`ele peut \^etre r\'esolu num\'eriquement gr\^ace \`a la formulation propos\'ee par J-F.Aujol \cite{aujol,aujol2}, en introduisant un param\^etre $\mu$ suppl\'ementaire correspondant \`a la norme maximale des textures dans l'espace $G$. L'utilisation des projecteurs non lin\'eaires d\'efinis par A.Chambolle \cite{chambolle} permet d'obtenir la d\'ecomposition de l'image par un algorithme it\'eratif (voir \cite{aujol,aujol2} pour tous les d\'etails).

\subsection{R\'ehaussement d'objets longilignes}
En imagerie a\'erienne ou satellitaire, une premi\`ere approximation est de voir les r\'eseaux routiers comme des objets filiformes. Lors d'exp\'erimentations sur la d\'ecomposition d'image, nous avons remarqu\'e que les structures filiformes \'etaient ``r\'ehauss\'ees'' dans la composante texture. Cette propri\'et\'e se justifie gr\^ace au lemme \ref{lem:route1}.

\begin{lemma}
\label{lem:route1}
Soit $f$ la fonction indicatrice de l'ensemble $E_N$ d\'efini par
\begin{equation}
E_N=[0,1]\times[0,N]
\end{equation}
quand $N$ est grand $f$ correspondra \`a une route dans l'image. On a alors

\begin{equation}
\|f\|_G\in\left[\left(2+\frac{2}{N}\right)^{-1},\frac{1}{2}\right]
\end{equation}
\end{lemma}

\begin{proof}
Soit la fonction $\theta$ d\'efinie par
\begin{equation}
\theta (t)=
\begin{cases}
t-\frac{1}{2} \qquad &\text{si} \quad 0<t<1\\
\frac{1}{2} \qquad &\text{si} \quad t\geqslant 1\\
-\frac{1}{2} \qquad &\text{si} \quad t\leqslant 0\\
\end{cases}
\end{equation}
alors on peut \'ecrire la fonction $f$ comme la divergence de $\theta$ (afin respecter la d\'efinition de la norme sur l'espace $G$):
\begin{equation}
f=\chi_{E_N}=\frac{\partial}{\partial x_1}\left(\theta(x_1)\chi_{[0,N]}(x_2)\right)
\end{equation}
on a alors
\begin{equation}
\left\|\theta(x_1)\chi_{[0,N]}(x_2)\right\|_{L^{\infty}}=\frac{1}{2}
\end{equation}
Ce qui nous fournit la borne sup\'erieure. La borne inf\'erieure est obtenue en appliquant la propri\'et\'e suivante
\begin{equation}
\|f\|_G\|f\|_{BV}\geqslant \int f^2(x)dx=|E_N|
\end{equation}
On en d\'eduit
\begin{equation}
\|f\|_G\geqslant \frac{N}{2N+2}
\end{equation}
ce qui ach\`eve la d\'emonstration.
\end{proof}

\begin{figure}[ht!]
\begin{center}
\includegraphics[scale=0.35]{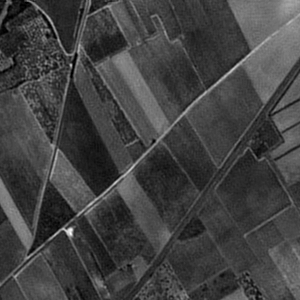}
\includegraphics[scale=0.35]{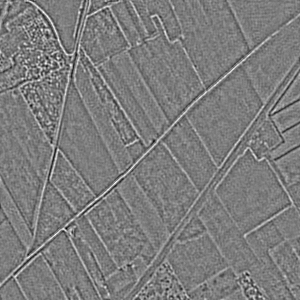}
\includegraphics[scale=0.35]{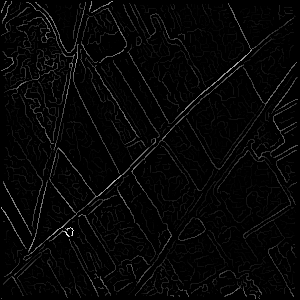}
\caption{R\'ehaussement d'objets longilignes: \`a gauche l'image originale, au milieu la composante texture issue de la d\'ecomposition dans laquelle les routes sont r\'ehauss\'ees et \`a droite le r\'esultat en sortie du filtre de Canny-Deriche.}
\label{fig:reseau}
\end{center}
\end{figure}

La figure~\ref{fig:reseau} illustre cette propri\'et\'e sur une imagette extraite d'une image a\'erienne. La d\'ecomposition peut donc servir de pr\'etraitement avant une \'etape de d\'etection \`a proprement parler. Afin de v\'erifier la plus-value de cet espace de rep\'esentation, nous utilisons un algorithme de bas niveau construit \`a partir d'une d\'etection d'alignements et d'un raffinement par contour actifs d\'ecris dans les sections suivantes.

\section{D\'etection d'alignements}
La composante texture de la figure~\ref{fig:reseau} nous montre qu'en premi\`ere approximation nous pouvons consid\'erer une route comme \'etant un alignement de points. Un moyen simple de d\'etecter ce type de configuration g\'eom\'etrique est le d\'etecteur d'alignements issu de la th\'eorie de la \textit{Gestalt} propos\'e par Morel et al. (voir \cite{gestalt1} pour tous les d\'etails).
Afin d'\'eliminer certaines redondances de segments d\'etect\'es mais correspondant \`a une m\^eme route, nous incorporons quelques r\`egles de ``fusion'' entre les segments (ces r\`egles sont disponibles dans \cite{jegilles2}).
L'image de gauche de la figure~\ref{fig:resa1} illustre le r\'esultat obtenu sur la partie texture de la figure~\ref{fig:reseau}. Cet algorithme ne fournissant que des segments, une \'etape suppl\'ementaire est n\'ecessaire afin de retrouver les courbes r\'eelles du r\'eseau routier. Cette \'etape est d\'ecrite dans la section suivante.
\section{Raffinement par contours actifs}

\begin{figure}
\centering\includegraphics[scale=0.4]{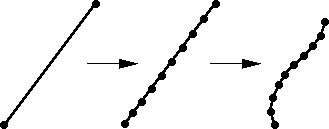}
\caption{Principe de conversion des segments en contours actifs.}
\label{fig:snakec}
\end{figure}

\begin{figure}
\begin{center}
\includegraphics[scale=0.35]{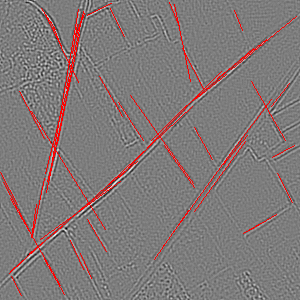}
\includegraphics[scale=0.35]{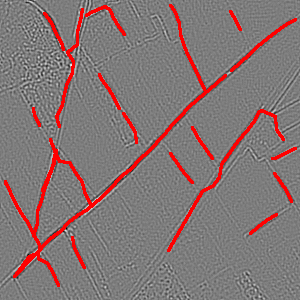}
\caption{A gauche: r\'esultats en sortie de l'algorithme de d\'etection d'alignements. A droite, r\'eseau extrait apr\`es raffinement par contours actifs.}
\label{fig:resa1}
\end{center}
\end{figure}

L'algorithme de d\'etection d'alignements ne fournit qu'une liste de segments. Ceux-ci ne refl\`etent donc pas les formes r\'eelles des routes. Afin de r\'esoudre ce probl\`eme, nous proposons de convertir chaque segment d\'etect\'e en une courbe polygonale ouverte compos\'ee de n\oe uds r\'eguli\`erement espac\'es (voir la figure~\ref{fig:snakec}). Etant donn\'e que les courbes initiales sont proches des routes recherch\'ees, nous pouvons appliquer le principe d'\'evolution des courbes d\'ecrit dans \cite{jegilles} (le principe \'etant de faire \'evoluer les courbes suivant la normale en chaque n\oe ud par un processus de recherche de la meilleure position).\\
Cet algorithme appliqu\'e sur les segments pr\'ec\'edemment d\'etect\'es fournit le r\'esultat illustr\'e sur l'image de droite de la figure~\ref{fig:resa1}. Nous voyons que la topologie des routes est bien retrouv\'ee gr\^ace aux contours actifs.

\section{Exp\'erimentation}
\begin{figure}
\begin{center}
\includegraphics[scale=0.35]{snakes}
\includegraphics[scale=0.35]{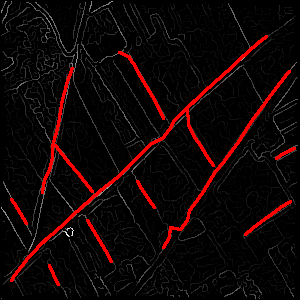}
\caption{A gauche: r\'esultats obtenus \`a partir de la composante texture. A droite, r\'eseau extrait \`a partir d'un filtrage de Canny-Deriche.}
\label{fig:deriche}
\end{center}
\end{figure}

Afin d'\'evaluer l'int\'er\^et de l'utilisation de l'espace des textures comme espace de r\'epr\'esentation, nous comparons les r\'esultats en sortie de notre d\'etecteur obtenus \`a partir de cet espace et d'une image pr\'etrait\'ee par une d\'etection de contours classique (Canny-Deriche). La figure~\ref{fig:deriche} donne les deux r\'esultats obtenus. Nous avons pu v\'erifier sur plusieurs images que la d\'etection par l'utilisation de l'espace des textures donne de meilleurs r\'esultats.

\section{Conclusion}
\begin{figure}[h!]
\centering\includegraphics[scale=0.15]{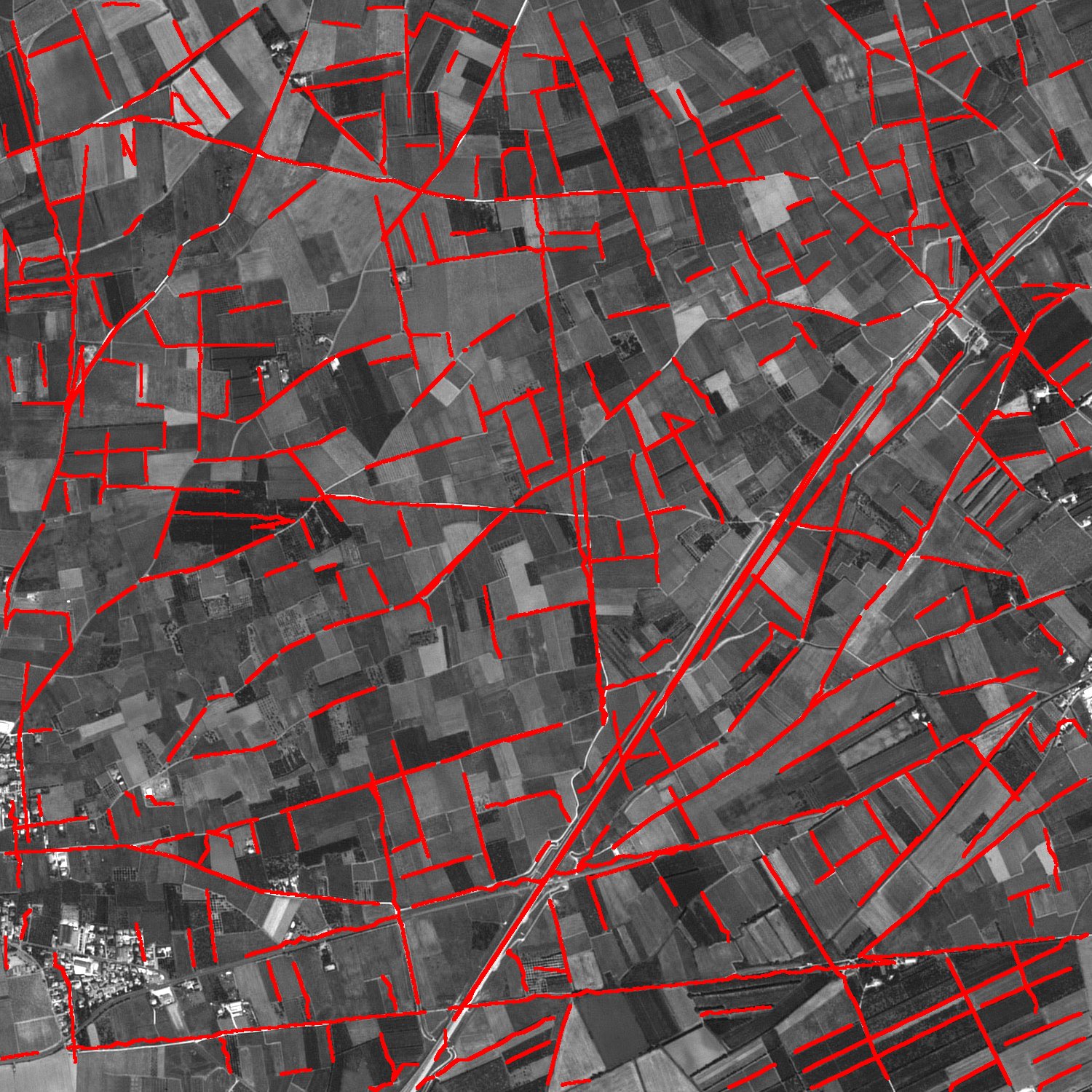}
\caption{D\'etection obtenue sur une image 4000x4000 pixels.}
\label{fig:snake}
\end{figure}

Dans cet article, nous montrons qu'une d\'ecomposition de l'image dans l'espace des textures permet de r\'ehausser sp\'ecifiquement les objets filiformes comme les r\'eseaux routiers.\\ 
Les premiers r\'esultats obtenus sur des images de taille beaucoup plus cons\'equente sem\-blent tr\`es encourageants (voir la figure~\ref{fig:snake}). Un algorithme de d\'etection de plus haut niveau permettra de discriminer les routes des autres objets longilignes (comme les bords des parcelles de champs) comme par exemple les travaux de Rochery et al. (\cite{rochery}). \\
Des tests sur une base de donn\'ee plus cons\'equente et disposant des v\'erit\'es terrains permettrait d'obtenir une \'evaluation plus pouss\'ee de l'algorithme. Le choix des param\^etres, notamment en ce qui concerne l'algorithme de d\'ecomposition ($\lambda$ et $\mu$), est en cours d'\'etude dans le cadre d'un travail plus th\'eorique sur les m\'ethodes de d\'ecomposition elles-m\^emes.\\
Une autre \'evolution int\'eressante \`a \'etudier est l'extension de l'algorithme au cas des images de type SAR. En effet, ce type d'image \'etant particuli\`erement bruit\'e, ce bruit est extrait dans la composante texture alt\'erant ainsi la d\'etection d'alignements. Une piste envisag\'ee est l'extension des m\'ethodes de d\'ecomposition \`a trois composantes (structures+textures+bruit) propos\'ees dans \cite{jegilles2} au cas du bruit multiplicatif.

\bibliographystyle{plain}
\bibliography{taima07}

\end{document}